\documentclass[10pt,twocolumn,letterpaper]{article}

\usepackage[pagenumbers]{cvpr} 

\usepackage{graphicx}
\usepackage{multirow}
\usepackage{placeins}
\usepackage{cuted}
\usepackage[misc]{ifsym}

\makeatletter
\renewcommand\paragraph{%
  \@startsection{paragraph}{4}{\z@}%
    {1ex \@plus 0.2ex \@minus 0.2ex}
    {-1em}
    {\normalfont\normalsize\bfseries}%
}%
\makeatother

\definecolor{cvprblue}{rgb}{0.21,0.49,0.74}
\usepackage[pagebackref,breaklinks,colorlinks,allcolors=cvprblue]{hyperref}

\def\paperID{11556} 
\def\confName{CVPR}
\def\confYear{2026}

\title{X-Humanoid: Robotize Human Videos to Generate Humanoid Videos at Scale}

\author{Pei Yang\thanks{Equal contribution.} \quad Hai Ci\footnotemark[1] \quad Yiren Song \quad Mike Zheng Shou\thanks{Corresponding author.}\\
\\
Show Lab, National University of Singapore\\
\\
{\small \url{https://showlab.github.io/X-Humanoid/}}
}

\begin{document}
\maketitle

\begin{strip}
\centering
    \includegraphics[width=0.98\linewidth]{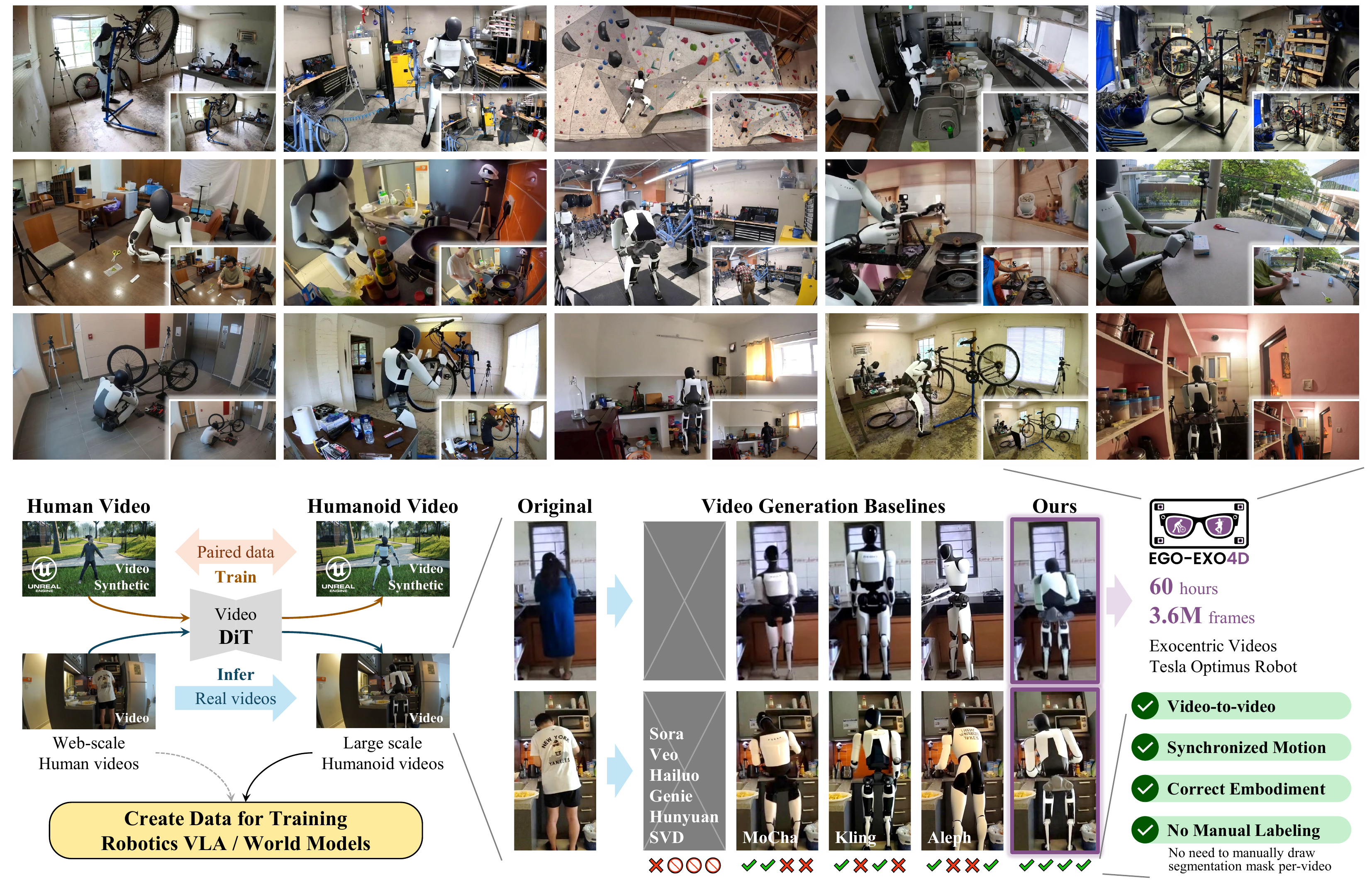}
    \captionof{figure}{Our proposed paradigm for ``robotizing'' human videos. \textbf{(1)} We adapt a Video Diffusion Transformer (DiT) into a video-to-video architecture. \textbf{(2)} This model is finetuned on a novel synthetic dataset of paired human-humanoid videos we generated using Unreal Engine. \textbf{(3)} At inference, the finetuned model translates real-world human videos (e.g., Ego-Exo4D \cite{egoexo4d}) into a large-scale dataset of humanoid videos, which can be used to train downstream robotics VLA or world models.}
    \label{fig:teaser}
\end{strip}

\begin{abstract}
    The advancement of embodied AI has unlocked significant potential for intelligent humanoid robots. However, progress in both Vision-Language-Action (VLA) models and world models is severely hampered by the scarcity of large-scale, diverse training data. A promising solution is to ``robotize" web-scale human videos, which has been proven effective for policy training. However, these solutions mainly ``overlay" robot arms to egocentric videos, which cannot handle complex full-body motions and scene occlusions in third-person videos, making them unsuitable for robotizing humans. To bridge this gap, we introduce X-Humanoid, a generative video editing approach that adapts the powerful Wan 2.2 model into a video-to-video structure and finetunes it for the human-to-humanoid translation task. This finetuning requires paired human-humanoid videos, so we designed a scalable data creation pipeline, turning community assets into 17+ hours of paired synthetic videos using Unreal Engine. We then apply our trained model to 60 hours of the Ego-Exo4D videos, generating and releasing a new large-scale dataset of over 3.6 million ``robotized" humanoid video frames. Quantitative analysis and user studies confirm our method's superiority over existing baselines: 69\% of users rated it best for motion consistency, and 62.1\% for embodiment correctness. 
\end{abstract}
    
\section{Introduction}

Recent advances in robotics, such as vision-language-action (VLA) models \cite{rt1, roboflamingo, pi0} and world models \cite{cosmos, cosmospredict25}, show emerging promise for general-purpose autonomy. However, current research is suffered from data scarcity \cite{immimic, masquerade}. Access to such large-scale robot data would enable more straightforward training and potentially unlock a higher performance upper bound.

A promising strategy to alleviate this data scarcity is to leverage the vast existing human activity videos \cite{phantom, masquerade, vitra}. Here, a significant challenge arises from the \textbf{visual embodiment gap}: the distinct physical appearance of humans versus robots prevents the direct use of this data. Some approaches attempt to ``robotize" human videos by editing the content of egocentric videos, such as replacing human arms with rendered robot arms \cite{masquerade, h2r}. While beneficial, these rule-based overlay methods often produce artifacts (e.g., wrong occlusions) and have not been successfully applied to third-person videos. This is because the third-person scenario is substantially more complex, involving full-body motions, dynamic backgrounds, and severe occlusions that are beyond the capabilities of simple inpaint-and-overlay techniques. To effectively transform third-person human videos into humanoid videos, and in turn generate large-scale data to alleviate data scarcity, we take a new approach: leveraging modern generative video models for flexible human-to-humanoid video editing.

We introduce X-Humanoid (X denotes ``transfer"), adapting the powerful Wan 2.2 \cite{wan} diffusion transformer (DiT) model into a video-in video-out structure. The model is then finetuned to take a human video as input, outputting a video where the person is replaced by a fixed humanoid embodiment. The humanoid's motion should be consistent with the original human's action, while all other video content should be best preserved, such as the background scene.

\begin{figure*}
    \centering
    \includegraphics[width=\linewidth]{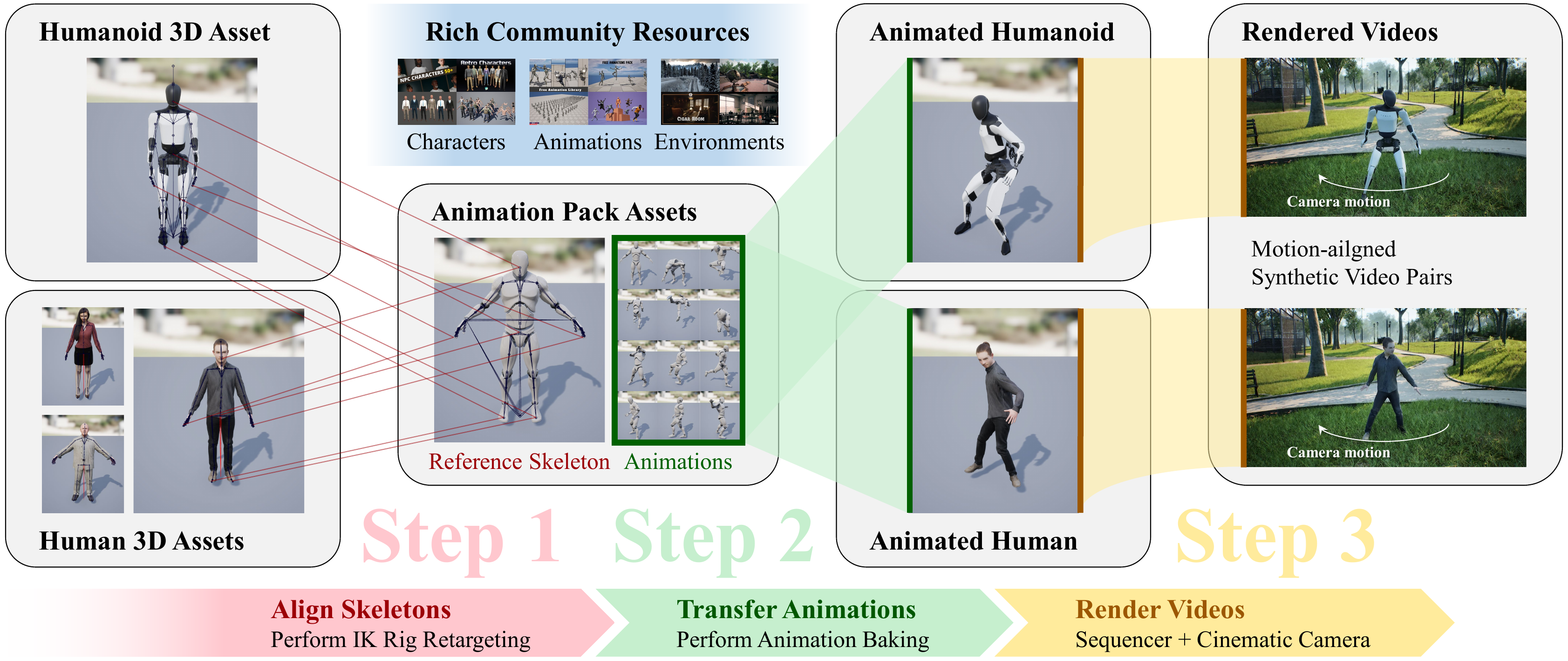}
    \caption{Our synthetic data creation pipeline. We leverage rich community assets (e.g., Fab marketplace) for characters, animations, and environments (blue) to create paired human-humanoid videos in three steps: First (red), we align the skeletons of different characters and animations to ensure compatibility across all assets. Second (green), with the help of compatible skeletons, we transfer the same animation to both human and humanoid characters. Finally (yellow), we place the animated human and humanoid in diverse scenes and record them using identical camera setups and movements to produce the paired data.}
    \label{fig:methodology_data}
\end{figure*}

Training this model requires paired human-humanoid videos, which are not yet available. We therefore designed a novel data synthesis method using Unreal Engine, turning rich community resources into paired human-humanoid videos performing identical animations. We generate over 17 hours of paired 1080p 30 fps video data (Fig. \ref{fig:methodology_data}). Using only 6.4\% of this new synthetic dataset, we performed LoRA finetuning on our adapted Wan 2.2 model. As illustrated in Fig.~\ref{fig:teaser}, this finetuned model can then ``robotize'' real-world web-scale videos. We apply our trained model to create a large-scale, 60-hour (3.6 million frames) ``robotized'' version of the Ego-Exo4D \cite{egoexo4d} dataset, featuring the Tesla Optimus humanoid.

We validate our approach through comparisons against existing video editing baselines. Both quantitative metrics and a user study with 29 participants confirm our method's superior performance. For instance, 69.0\% of users rated our results as having the best motion consistency, and 75.0\% preferred our method for overall video quality. Furthermore, by leveraging a powerful base model, our approach successfully edits challenging in-the-wild videos, faithfully preserving complex effects such as motion blur and adapting to the original video's quality.

Our main contributions are summarized as follows:

\begin{itemize}
    \item We introduce a generative video editing approach to address data scarcity in robotics research, by adapting and finetuning a modern video generation model.
    \item We introduce a scalable pipeline to synthesize paired human-humanoid videos in Unreal Engine, releasing a new 17+ hour, 1080p 30 fps dataset to train our model.
    \item We create and release a large-scale, 60+ hour ``robotized" dataset edited from Ego-Exo4D, featuring the Tesla Optimus humanoid, to help alleviate robotics data scarcity.
\end{itemize}

\section{Related Work}

\label{sec:related_works}

\begin{figure*}
    \centering
    \includegraphics[width=\linewidth]{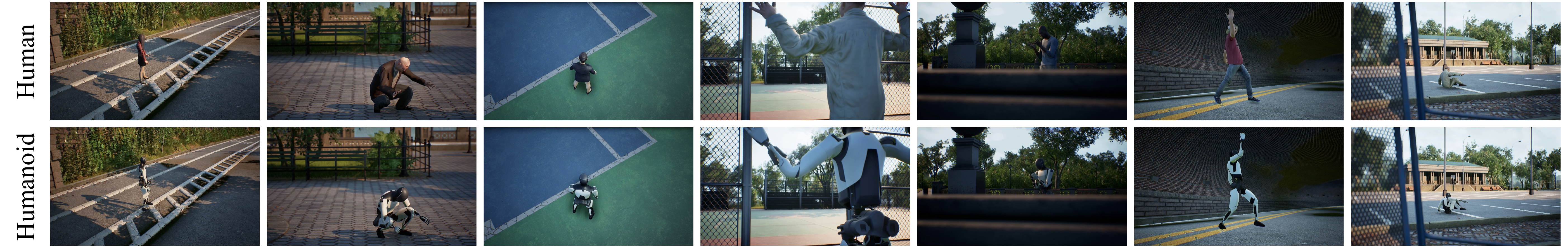}
    \caption{Visualization of sample pairs from our synthesized Human-Humanoid video dataset. To ensure diversity, the dataset encompasses a wide variety of scenes, character motions, and camera parameters (e.g., focal length, exposure). We intentionally include challenging conditions, such as occlusions by obstacles and partial-body or off-center framing, to improve model robustness.}
    \label{fig:visualise_synthetic}
\end{figure*}

\paragraph{Data Scarcity in Robotics Research} Current robotics research is advancing in two directions: (1) building robotic policies based on vision-language-action (VLA) models for instruction-following \cite{rt1, roboflamingo, leo, gr1, octo, openvla, tinyvla, pi0, uniact, openvla, gr00t}, and (2) training world models for future-state prediction \cite{worldmodels, dreamtocontrol, cosmos, cosmospredict25}. Both areas are fundamentally hindred by a scarcity of large-scale robot data, which limits policy generalization \cite{dreamgen} and future prediction accuracy \cite{aleph}. Leveraging abundant, existing videos of human activity from datasets \cite{ego4d, egoexo4d} or the Interent presents a significant opportunity to alleviate this data scarcity \cite{dreamgen, masquerade}. 

\paragraph{Leveraging Human Activity Videos} A significant challenge in utilizing human videos is the \textit{visual embodiment gap}: humans and robots possess distinct physical appearances, preventing the direct use of human videos for training. To circumvent this, prior works typically extract intermediate representations like poses or trajectories to guide execution \cite{videodex, dexvip, track2act}, or utilize specific reward functions to pretrain visual encoders \cite{egomimic, egozero, egovla, immimic}. While these methods successfully circumvented the \textit{visual domain gap} issue, a alternative strategy could be bridging the gap at the pixel level, by directly training robot policies or world models on robot videos \cite{masquerade}. With such videos, there is no \textit{visual embodiment gap} between the training videos and the actual robot, thus training could be more straightforward and potentially have a higher performance upper bound. Such data has even wider benefits, because aside from VLA-based policies, robot videos could also facilitate the training of world model-centered policies \cite{worldmodelphysical}. Unfortunately, manually collecting such data \cite{human2robot, twist2, phantom} is costly and limited in scale and scene diversity.

A more scalable alternative is to ``robotize" existing human videos by editing the video content to replace the human with a robot. \cite{masquerade, h2r, mimicdreamer, gigaworld0} have explored this for egocentric videos by inpainting the human arm and overlaying a rendered robot arm matching the human arm's pose or hand motion. Despite producing numerous artifacts (e.g., occlusion errors), this rule-based overlay data has been shown to improve VLA performance \cite{masquerade}. However, robotizing \textit{third-person} human videos, a task beneficial for both humanoid policies and world models, remains unexplored. This scenario is substantially more complex, involving intricate full-body motions, dynamic backgrounds, and severe occlusions that are beyond the capabilities of existing rule-based methods. To address this complex issue, in this work, we attempt to ``robotize" human videos by leveraging the capabilities of powerful, modern video generative models. 

\paragraph{Video Generation and Editing} Modern video generative models have evolved from simple text-to-video generation \cite{show1, tuneavideo} to complex systems handling mixed-modal inputs and outputs \cite{sora, gen3, genie3, minimax, veo3, moviegen, hunyuanvideo, i2vgenxl, pyramidflow}. Several existing models, such as Kling \cite{kling}, Aleph \cite{aleph}, VACE \cite{vace}, and MoCha \cite{mocha}, support image-conditioned video editing. These methods allows taking an input video (e.g., of a human) and a reference input image (e.g., of a robot) to produce an output video where the human is replaced by the robot. However, we found these models generally prone to poor editing results for our specific task. Therefore, we do not use them. Instead, we leverage the powerful Wan 2.2 \cite{wan} model, which we adapt into a video-to-video generation framework to effectively robotize human videos.

\section{Methodology}

\begin{figure*}
    \centering
    \includegraphics[width=0.92\linewidth]{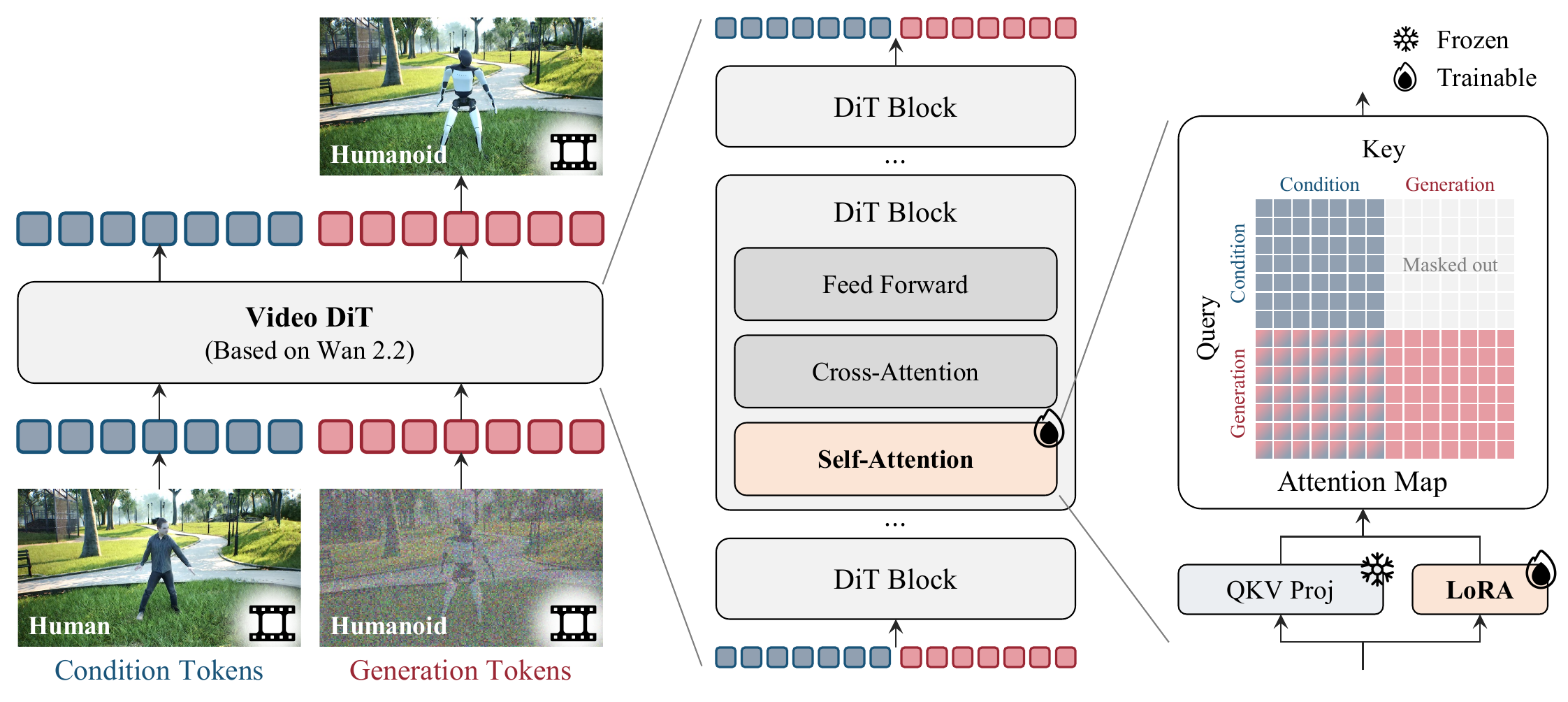}
    \caption{The network architecture of our method. The original Wan 2.2 model does not accept video input. It only contains generation tokens (red), which the model denoises to generate a video. Parallel to these generation tokens, we encode the input human video into condition tokens (blue) and concatenate all tokens. During self-attention, we apply a mask to prevent condition tokens from attending to generation tokens. At the output, we only retain the generation tokens and decode them to produce the edited video.} 
    
    \label{fig:methodology_model}
\end{figure*}

\subsection{Problem Formulation}

\label{sec:problem_formulation}

We aim to train a video-to-video generation model that performs \textit{third-person} human-to-humanoid editing. The model's input is a video of a human performing an activity, and its output is a video of a specific humanoid robot model doing the same activity within the identical scene. A critical constraint is that the robot's motion must be consistent with the human's action and strictly per-frame aligned.

To achieve this, we need to create paired training data (Sec. \ref{sec:methodology_data}), design a video-in video-out model architecture (Sec. \ref{sec:methodology_architecture}), and finetune the model (Sec. \ref{sec:methodology_finetuning}). We then use our model to ``robotize" more than 60 hours of Ego-Exo4D \cite{egoexo4d} videos, and also demonstrate the model's capabilities on editing diverse in-the-wild Internet videos.

\subsection{Creating Human-Humanoid Paired Videos}

\label{sec:methodology_data}

The video editing model's core capability is transforming the protagonist human into a humanoid with identical actions, keeping other scene details unchanged. We thus use Unreal Engine to synthesize paired videos of humans and humanoid performing the same actions in the same scene, using rich community-sourced assets.

Creating these paired videos involves three steps (Fig. \ref{fig:methodology_data}). First, to apply the same motion to different characters, we must resolve the skeletal incompatibility between our various character and animation assets. This is achieved through aligning the skeletons by manual \textit{IK Rig Retargeting}. Second, with the skeletons mapped, we transfer (\textit{a.k.a. bake}) each single motion animation to all human and humanoid characters. Third, we place both characters in the same scene, play the same animations, and record them with virtual cameras to get paired movie clips. To diversify the videos, we captured along various camera paths, intentionally capturing occlusions (e.g., behind obstacles) and off-center characters. These videos cover varied exposure and color grading settings, 14-80mm focal lengths, and \textit{f}/2.8-\textit{f}/5.6 apertures. 

In total, we synthesized 11172 pairs of 1080p 30fps videos across 14 scenes, containing a total of over 2.8 million frames. This took 10 days to render on one NVIDIA RTX 3060 GPU using Unreal Engine 5.3.2. Fig. \ref{fig:visualise_synthetic} visualizes sample paired data.

\begin{figure*}
    \centering
    \includegraphics[width=0.95\linewidth]{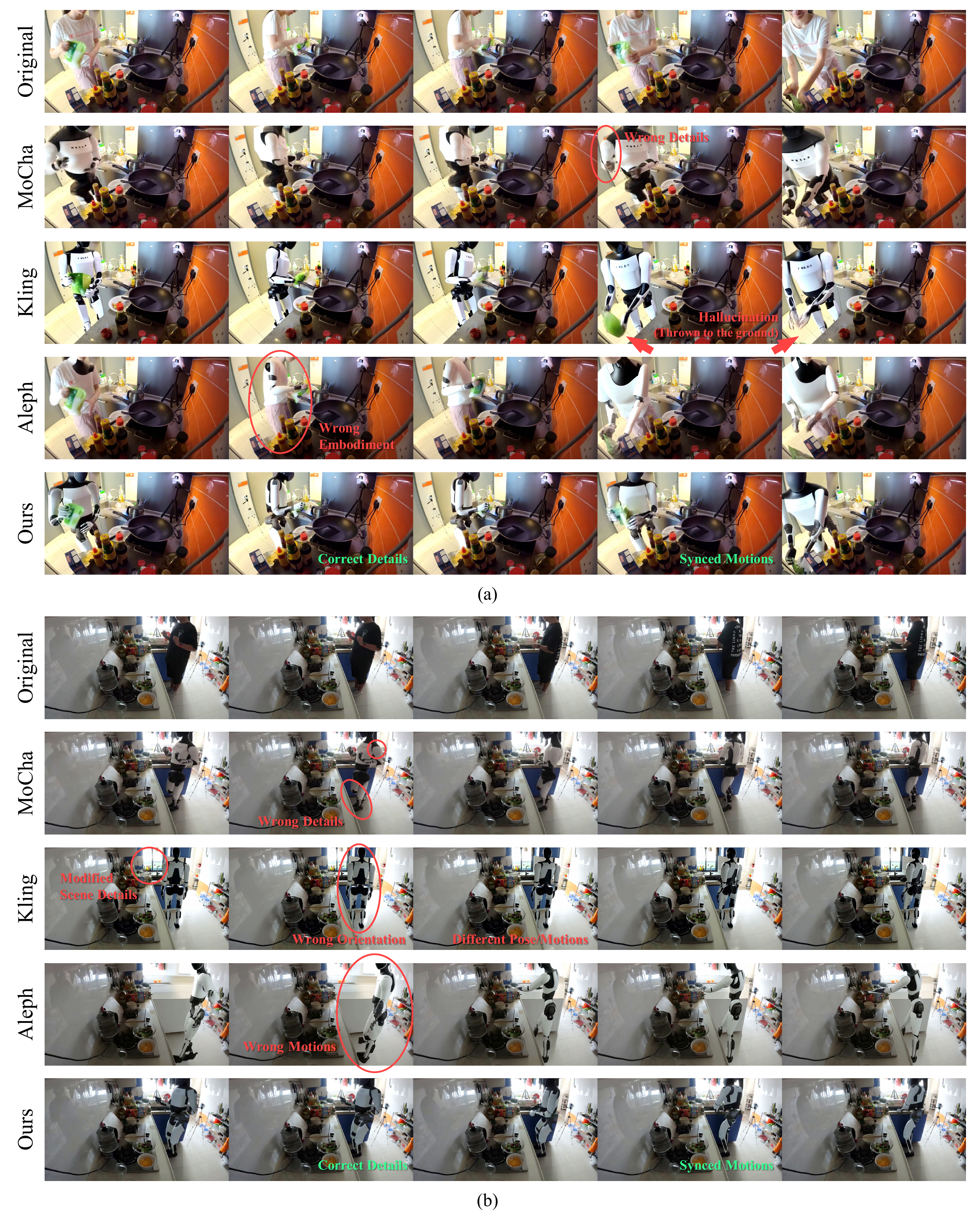}
    \vspace{-0.25cm}
    \caption{Qualitative comparison to baselines. The videos are sampled at frames 15, 30, 45, 60, and 75. MoCha \cite{mocha} generally maintains the original character's motion, but it often produces incorrect details, such as the upper arm shape in (a) and the leg pose in (b). Kling \cite{kling} generates the correct embodiment but fails on motion consistency, resulting in unsynchronized actions and undesirable scene alternations (e.g., dropping the green bag to the ground instead of putting it back in (a)). Runway Aleph \cite{aleph} fails to produce both the correct robot embodiment (a) and the correct motion (b). In contrasts, our method successfully generates the correct embodiment and motions synchronized with the original human.} 
    \label{fig:baseline_comparison}
\end{figure*}

\begin{table*}[]
\centering
\caption{Quantitative baseline comparison. This table presents a two-part evaluation. The first set of columns (\textbf{Similarity Metrics}) measures performance on our synthetic videos, where ground-truth data is available; we report PSNR, SSIM, and MSE (computed on pixel values [0, 255]). The second set (\textbf{User Preference Rate}) evaluates performance on real-world Ego-Exo4D \cite{egoexo4d} videos, which lack ground truth. Here, we report the percentage of times users preferred a method across four criteria, allowing multiple bests. The results show our method, which does not require manual per-video labeling, significantly outperforms all baselines across every metric.}
\label{tab:baseline_wide}
\resizebox{\textwidth}{!}{%
\begin{tabular}{lclcccccccc}
\hline
\multicolumn{1}{c}{\multirow[b]{2}{*}{\textbf{Method}}} & \multirow{2}{*}{\textbf{\begin{tabular}[c]{@{}c@{}}Manual\\ Per-Video\\ Label\end{tabular}}} & \multicolumn{1}{c}{} & \multicolumn{3}{c}{\textbf{Similarity Metrics (Synthetic Videos)}}                                                                                    &           & \multicolumn{4}{c}{\textbf{User Preference Rate (Ego-Exo4D Videos)}}                                                                                                                                                                                                                                       \\ \cline{4-6} \cline{8-11} 
\multicolumn{1}{c}{}                                 &                                                                                              & \multicolumn{1}{c}{} & \textbf{\begin{tabular}[c]{@{}c@{}}PSNR $\uparrow$\\ (dB)\end{tabular}} & \textbf{SSIM $\uparrow$}  & \textbf{\begin{tabular}[c]{@{}c@{}}MSE $\downarrow$\\ ({[}0, 255{]})\end{tabular}} & \textbf{} & \textbf{\begin{tabular}[c]{@{}c@{}}Motion\\ Consistency\end{tabular}} & \textbf{\begin{tabular}[c]{@{}c@{}}Background\\ Consistency\end{tabular}} & \textbf{\begin{tabular}[c]{@{}c@{}}Embodiment\\ Consistency\end{tabular}} & \textbf{\begin{tabular}[c]{@{}c@{}}Video\\ Quality\end{tabular}} \\ \hline
\textbf{Kling \cite{kling}}                                       & Require                                                                                      &                      & 16.952                                                       & 0.288          & 1518.785                                                              &           & 17.2\%                                                                & 6.9\%                                                                     & 3.79\%                                                                    & 3.1\%                                                            \\
\textbf{MoCha \cite{mocha}}                                       & Require                                                                                      &                      & 17.163                                                       & 0.315          & 1534.145                                                              &           & 20.7\%                                                                & 24.1\%                                                                    & 20.7\%                                                                    & 13.8\%                                                           \\
\textbf{Runway Aleph \cite{aleph}}                                       & Not require                                                                                  &                      & 17.683                                                       & 0.402          & 1295.640                                                              &           & 0.0\%                                                                 & 10.3\%                                                                    & 3.4\%                                                                     & 3.4\%                                                            \\
\textbf{Ours}                                        & Not require                                                                                  &                      & \textbf{21.836}                                              & \textbf{0.671} & \textbf{459.302}                                                      & \textbf{} & \textbf{69.0\%}                                                       & \textbf{75.9\%}                                                           & \textbf{62.1\%}                                                           & \textbf{62.1\%}                                                  \\ \hline
\end{tabular}%
}
\end{table*}


\subsection{Model Architecture}

\label{sec:methodology_architecture}

We adapt Wan 2.2 \cite{wan}, a Diffusion Transformer- (DiT-) based video generation model, into a video-in video-out architecture. As shown in Fig. \ref{fig:methodology_model}, we first encode the input video into a sequence of \textit{condition tokens} and concatenate them before the \textit{generation tokens} that are subsequently denoised. To establish a strict spatio-temporal correspondence for video editing, we use identical positional embeddings for both sets of tokens. To prevent the input condition from being corrupted by the generation process, we apply a unidirectional mask in each DiT blocks's self attention map, preventing conditional tokens from attending to generation tokens. Finally, the model is finetuned to denoise only the generation tokens to produce the output video, while the model's output on the conditional tokens is ignored.

\subsection{Finetuning}

\label{sec:methodology_finetuning}

We adopt a flow-matching finetuning objective, consistent with Wan 2.2's training \cite{wan}. The goal is to train the model $v_\theta$ to predict a ground-truth velocity $v_t$. This velocity is defined along a simple, linear path $x_t$ that interpolates from pure noise $x_0 \sim \mathcal{N}(0, I)$ to the clean latent tokens $x_1$ \cite{rectifiedflow}: $x_t = tx_1 + (1-t)x_0$, where $t \sim [0, 1]$ is the timestep. The ground-truth velocity $v_t$ is the derivative of this path:

\begin{equation}
    v_t = \frac{dx_t}{dt} = x_1 - x_0.
\end{equation}

\noindent The model is then finetuned with a mean-squared error loss to predict this velocity, with loss function given by

\begin{equation}
    L = \mathbb{E}_{x_1^\text{con}, x_t^\text{gen}, t} \lVert v_\theta(x_1^\text{con}, x_t^\text{gen}, c_\text{text}, t) - v_t  \rVert^2.
\end{equation}

\noindent Here, $x^\text{con}$ and $x^\text{gen}$ denote condition and generation tokens, and $c_\text{text}$ is a fixed conditional text embedding (``Humanoid video"). At inference, the model's predicted velocity is integrated from $t=0$ to $1$ to produce the final, edited video:

\begin{equation}
    x_1^\text{gen} = x_0 + \int_0^1 v_\theta(x_1^\text{con}, x_t^\text{gen}, c_\text{text}, t) dt.
\end{equation}

\section{Experiments}

\subsection{Experimental Setup}

\label{sec:experimental_setup}

\paragraph{Data Processing} We partition the synthetic video pairs from 14 virtual scenes, using 12 scenes for training and reserving 2 for validation. Both synthetic and real (Ego-Exo4D \cite{egoexo4d} and Internet) videos are downsampled to 15 fps, resized to 864$\times$400, and segmented into 90-frame (6-second) clips. Any clips shorter than 90 frames are discarded. For evaluation, we sample 50 clips each from this validation set and the Ego-Exo4D dataset.

\paragraph{Model Finetuning} We performed rank-96 LoRA finetuning on a \texttt{Wan2.2-TI2V-5B} model for 500 steps. We used four NVIDIA H200 GPUs using Distributed Data Parallel (DDP) with a batch size of 1 per GPU. We employ the AdamW optimizer with a learning rate of 1e-4, a 50-step linear warmup, a weight decay of 1e-2, and betas of 0.9 and 0.95. This finetuning process completed in 2.5 hours and consumed approximately 6.4\% of the synthetic data. 

\paragraph{Evaluation and Metrics} We conduct quantitative evaluations using a two-pronged approach. Primarily, to assess performance on real-world Ego-Exo4D \cite{egoexo4d} human videos (which lack ground truth), we conduct user studies. 29 participants with computer vision or robotics backgrounds compared our method against baselines, evaluating 10 clips each. They selected the best-performing method(s) across four criteria: \textbf{motion consistency}, \textbf{background consistency}, \textbf{embodiment consistency} (similarity to the Tesla Optimus humanoid), and \textbf{overall video quality}. We report the \textbf{preference rate} (the percentage of times a method was chosen as best). Secondly, on our synthetic validation set where ground-truth videos are available, we quantitatively measure similarity using \textbf{PSNR}, \textbf{SSIM}, and \textbf{MSE}.

\subsection{Baseline Comparison}

\paragraph{Baseline Methods} We adopt Aleph \cite{aleph}, Kling \cite{kling}, and MoCha \cite{mocha} as our baseline methods. All are image-conditioned video editing models or systems (as introduced in Sec.~\ref{sec:related_works}). Kling and MoCha additionally require a manually labeled mask image (created with the help of SAM \cite{sam}), specifying the character to edit.

\paragraph{Our method best preserves the original character’s motion.} Quantitatively, as shown in Tab.~\ref{tab:baseline_wide}, 69.0\% of users considered our results to have the best or joint-best motion consistency with the original video. Qualitatively, our method demonstrates better motion consistency than MoCha, as evidenced by the leg poses in Fig.~\ref{fig:baseline_comparison}(a) and the hand movement when grasping the green package in Fig.~\ref{fig:baseline_comparison}(b). In both figures, Kling and Aleph produces noticeably inconsistent and motions from the original video: even the robot’s position and orientation could be different.

\paragraph{Our method best preserves the robot embodiment’s appearance.} As shown in Tab.~\ref{tab:baseline_wide}, 62.1\% of users rated our generated robot as most consistent (or tied for most consistent) with Tesla Optimus' appearance, followed by Kling at 37.9\%. Qualitatively, our method accurately reproduces key details of the robot’s shoulder and leg joints, which baseline methods more oftenly fail to capture. According to survey feedback, the main factor affecting user preference for our results is the (face) mask material, as our method renders the reflective mask as matte. This is because our synthetic videos exhibits such matte masks (as shown in Fig.~\ref{fig:visualise_synthetic}), and the model has faithfully learned this feature.

\paragraph{Our method best preserves other aspects of the video content, achieving overall superior quality.} Around three quarters of users preferred our video for background consistency and overall quality. As shown in Fig.~\ref{fig:baseline_comparison}, both our method and MoCha largely maintain environmental consistency with the original video. Kling, however, introduces additional elements (such as a faucet and a window in Fig.~\ref{fig:baseline_comparison}(a)), indicating a tendency toward video reinterpretation rather than faithful preservation of motion and details. While such reinterpretation can sometimes reduce artifacts and improve perceived quality, it deviates from the intended goal of “robotizing” videos, and the generated results may not be physically suitable for VLA or world model training. Supp. \ref{sec:appendix_baseline} contains all test set videos robotized by our methods, which are randomly sampled. From them, one can see the usability rate in their own use cases.

\begin{figure}
    \centering
    \includegraphics[width=\linewidth]{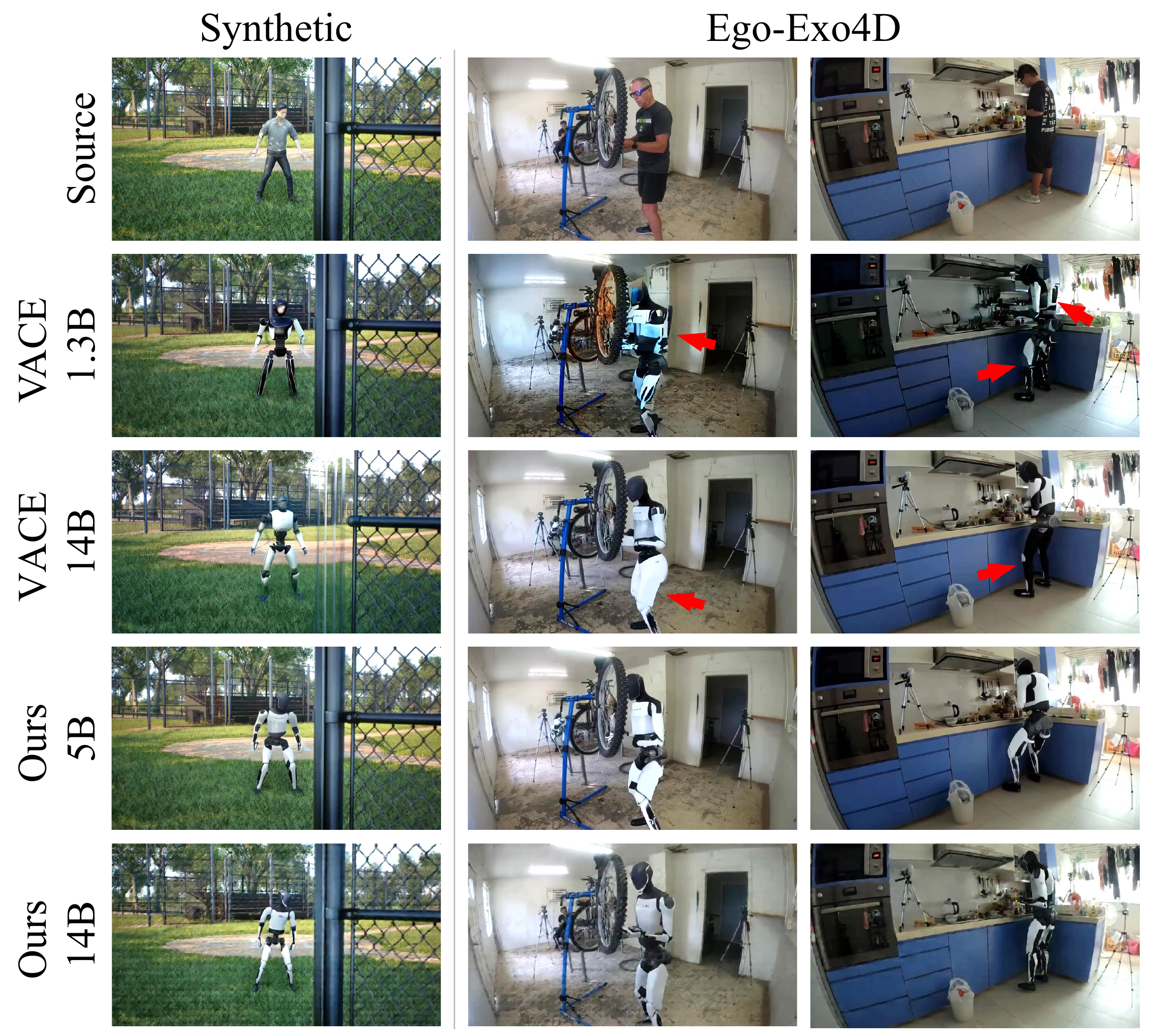}
    \caption{Qualitative ablation of model architectures. Both VACE models fail to generate the correct robot embodiment, while the 14B version of our model suffers from lower overall video quality. Our 5B model performs the best on this video editing task.} 
    \label{fig:ablation_model}
\end{figure}

\begin{figure}
    \centering
    \includegraphics[width=0.95\linewidth]{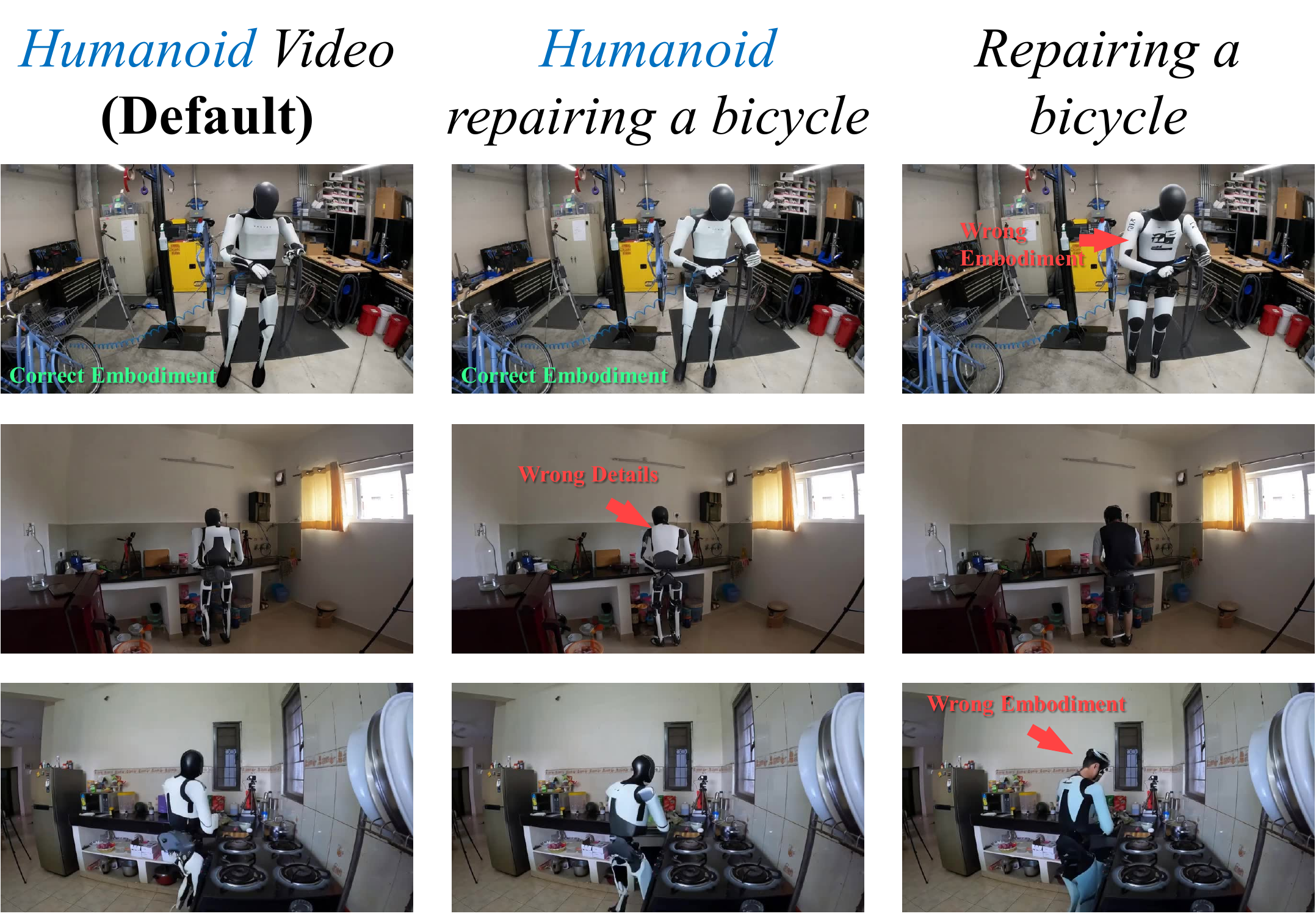}
    \caption{Qualitative ablation on the inference prompt. The finetuning prompt (left) performs best. Any other prompts could lead to sub-optimal performances.} 
    \label{fig:ablation_prompts}
\end{figure}

\begin{figure*}
    \centering
    \includegraphics[width=\linewidth]{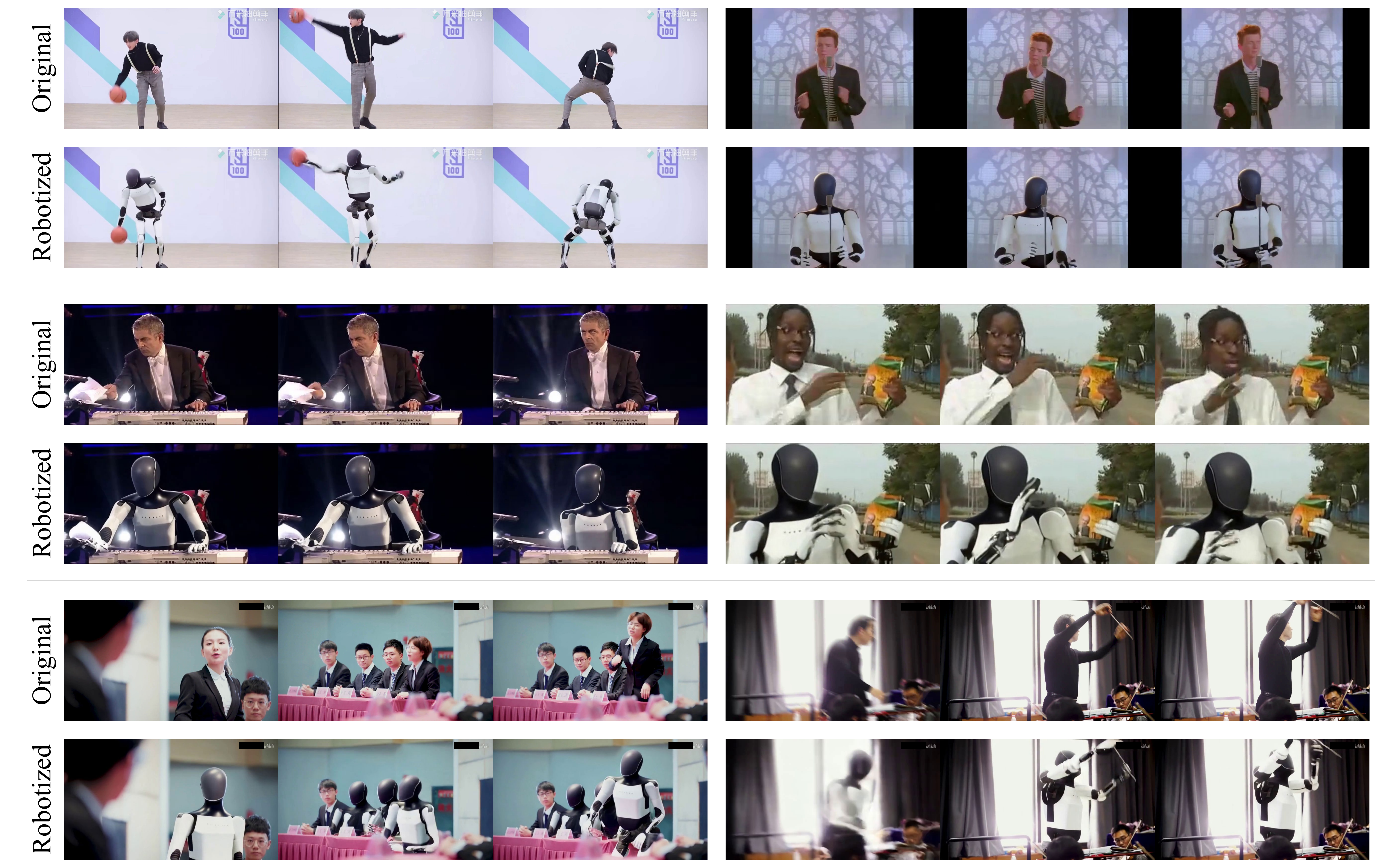}
    \caption{Visualization of robotizing in-the-wild videos. Our method successfully transforms the human protagonist into the correct humanoid embodiment while preserving motion, background, and overall video quality (e.g., middle-left vs. middle-right). It also robustly handles complex video effects such as pillarboxes (top-right), camera cuts (bottom-left), and motion blur (middle-right and bottom-right). } 
    \label{fig:internet_video}
\end{figure*}

\begin{table}[]
\centering
\caption{Ablation on model architecture. Similarity metrics are measured between output and ground-truth videos on the validation set (synthetic). Our 5B model achieves the best performances.}
\label{tab:ablation_model}
\resizebox{\linewidth}{!}{%
\begin{tabular}{lcccc}
\hline
           & \textbf{VACE 1.3B} & \textbf{VACE 14B} & \textbf{Ours 5B} & \textbf{Ours 14B} \\ \hline
\textbf{PSNR $\uparrow$ (dB)}          & 16.424             & 18.018            & \textbf{21.836}  & 21.311            \\
\textbf{SSIM $\uparrow$}               & 0.265              & 0.349             & \textbf{0.671}   & 0.598             \\
\textbf{MSE $\downarrow$ ({[}0, 255{]})} & 1906.218           & 1240.809          & \textbf{459.302} & 527.561           \\ \hline
\end{tabular}%
}
\end{table}

\begin{table}[]
\centering
\caption{Computational costs for our 5B and 14B models, normalized per video frame (batch size is 1). The 14B model's training and inference times are over 10 times greater than the 5B model's, making it unsuitable for efficiently robotizing web-scale videos.}
\label{tab:time}
\resizebox{0.88\linewidth}{!}{%
\begin{tabular}{lcc}
\hline
                                        & \textbf{Ours 5B} & \textbf{Ours 14B} \\ \hline
\textbf{Training Time (s/video frame)}  & \textbf{0.10}    & 1.05              \\
\textbf{Inference Time (s/video frame)} & \textbf{5.00}    & 69.16             \\ \hline
\end{tabular}%
}
\end{table}

\begin{table}[]
\centering
\caption{Ablation on finetuning steps. Similarity metrics are measured between output and ground-truth videos on the validation set (synthetic). Although more steps leads to better similarity, this reflects overfitting to the synthetic data, as shown in Fig. \ref{fig:ablation_steps}.}
\label{tab:ablation_steps}
\resizebox{0.85\linewidth}{!}{%
\begin{tabular}{lccc}
\hline
& \textbf{500 Steps} & \textbf{1500 Steps} & \textbf{2500 Steps}     \\ \hline
\textbf{PSNR $\uparrow$ (dB)}          & 21.836       & 21.989      & \textbf{22.264} \\
\textbf{SSIM $\uparrow$}               & 0.671        & 0.664      & \textbf{0.665} \\
\textbf{MSE $\downarrow$ ({[}0, 255{]})} & 459.302      & 439.360      & \textbf{415.034} \\ \hline
\end{tabular}%
}
\end{table}

\subsection{Ablation Study}

\label{sec:ablation_study}

\paragraph{Ablation on model choice and architecture.} We also tested our method on \texttt{Wan2.2-T2V-A14B}, as well as two VACE \cite{vace} models \texttt{Wan2.1-VACE-1.3B} and \texttt{Wan2.1-VACE-14B} (details in Supp. \ref{sec:appendix_vace}), because these VACE models also support video-in video-out. As shown in Fig.~\ref{fig:ablation_model}, neither VACE model was able to generate a correct robot embodiment, while our 14B model produced lower visual quality than the 5B version. This is also supported by quantitative evaluations on the validation set (Tab.~\ref{tab:ablation_model}). Moreover, the 14B model increased computation time by at least tenfold (Tab.~\ref{tab:time}). Therefore, our method based on \texttt{Wan2.2-TI2V-5B} represents the optimal choice.

\paragraph{Ablation on finetuning steps.} We finetune our model for 500 steps for an optimal performance. With too few steps (e.g., 200), the robot appears to be simply overlaid onto the video frame, lacking correct occlusion with the environment, and the embodiment itself looks less realistic (see Fig.~\ref{fig:ablation_steps} in Supp. \ref{sec:appendix_steps}). Conversely, with longer finetuning (e.g., 700 steps), the model begins to overfit to the synthetic data's domain. Its performance on the validation set continues to improve (Tab.~\ref{tab:ablation_steps}), but starts to become worse on real videos, resulting in clearly unreasonable content (such as fusing the bicycle's frame with the humanoid's leg, as shown in Fig.~\ref{fig:ablation_steps}).

\paragraph{Ablation on how text prompt affects editing performance.} As compared in Fig.~\ref{fig:ablation_prompts}, using the original finetuning prompt (``Humanoid video") achieves the best performance. If using other prompts, the model tends to ignore the text descriptions and prioritizes following the video's original motion, except for the keyword ``Humanoid". Removing this keyword (right column) leads to generating incorrect robot embodiments, indicating the model has associated this keyword with the robot's appearance.

\subsection{Robotizing In-the-wild Internet Videos} 

Finally, we apply our method to robotize diverse Internet videos. As shown in Fig.~\ref{fig:internet_video}, our method seamlessly replaces the protagonist while preserving the original motion (top-left), fine details (middle-left), camera cuts (bottom-left), aspect ratio (top-right), video quality (middle-right), and even complex effects like motion blur (middle- and bottom-right). We attribute these capabilities to the preservation of Wan 2.2's powerful generative capability. The model's success across these diverse scenarios demonstrates that our video editing approach can effectively robotize web-scale in-the-wild videos, potentially generating substantial data for training robot policies and world models.


\section{Conclusion}

In this work, to address the data scarcity problem, we studied how to generatively edit human videos into humanoid videos. The editing was achieved by a video-to-video model finetuned on self-created synthetic video pairs. By releasing both the 17-hour synthetic dataset and a 60-hour robotized version of Ego-Exo4D, we provide valuable new resources for research in humanoid VLA and world models.

\paragraph{Limitations and Future Analysis} As a pioneering work to robotize exocentric human videos, we mainly consider single-person videos, leading to undefined behavior in multi-person scenes (e.g., Fig. \ref{fig:internet_video}, bottom-left); future work could address this by adding explicit controls. Our method also requires training new LoRAs for new humanoid embodiments, while future work could extend to one-shot, image-conditioned methods.

\FloatBarrier

{
    \small
    \bibliographystyle{ieeenat_fullname}
    \bibliography{main}
}

\clearpage
\setcounter{page}{1}
\maketitlesupplementary

\setcounter{section}{0}
\renewcommand{\thesection}{\Alph{section}}

\section{Implementation Details of VACE Models}

\label{sec:appendix_vace}

As part of our ablation study in Sec. \ref{sec:ablation_study}, we included two VACE \cite{vace} models: \texttt{Wan2.1-VACE-1.3B} and \texttt{Wan2.1-VACE-14B}, both of which utilized \texttt{vace\_video} as an extra input to accept the source human videos. We adopted the DiffSynth implementation and trained these models on the identical dataset and partitions as our main method. We employed LoRA finetuning, applying it to the QKV and output projections in all attention layers, as well as the first and third layers in all feed-forward networks. Both models were trained using the AdamW optimizer with a learning rate of 1e-4. The 1.3B model was trained for 2500 steps over 14 hours, and the 14B model was trained for 16 hours, both using DDP on 4 H200 GPUs. Notably, we initially attempted to use the same fixed "Humanoid video" prompt as our main method, but this led to numerous failure cases. We therefore switched to using Qwen2.5-VL (\texttt{Qwen2.5-VL-7B-Instruct}) \cite{qwen25vl} to summarize each video, using this summary as the text prompt, which significantly reduced failures.

\section{Qualitative Ablation on Finetuning Steps}

\label{sec:appendix_steps}

\begin{figure}
    \centering
    \includegraphics[width=\linewidth]{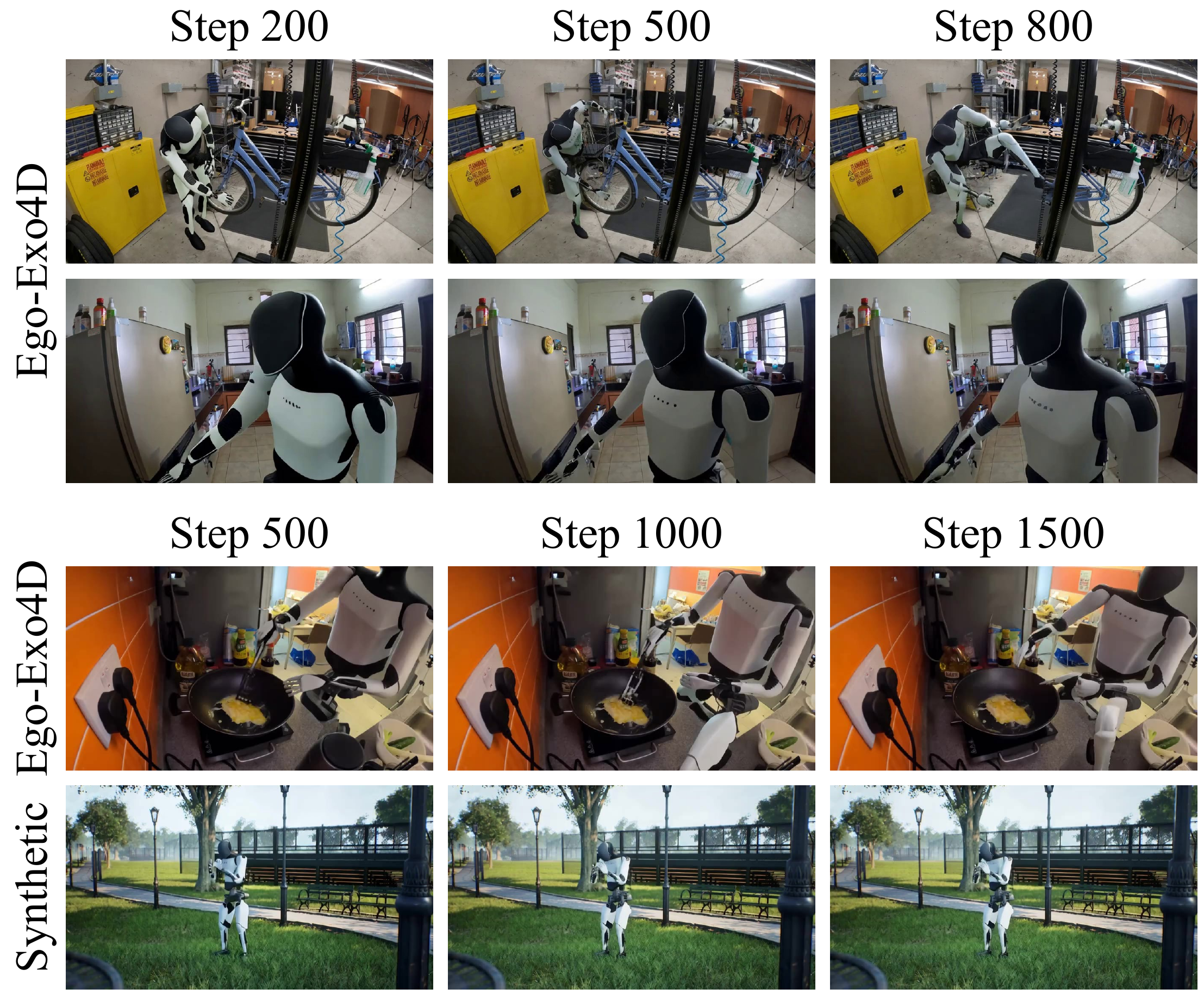}
    \caption{Qualitative ablation on finetuning steps. Too few steps (200) leads to incorrect occlusions (robot ``overlayed" in front of the bicycle), while more steps (700) leads to overfitting to the synthetic data, causing degraded generation when editing real videos.} 
    \label{fig:ablation_steps}
\end{figure}

Fig. \ref{fig:ablation_steps} shows our model's editing performance when finetuned with different steps. With too few steps (200), the model fails to generate with correct occlusions, as it ``overlays'' the robot in front of the objects, such as the bicycle. With more steps (700), the model starts to overfit to the synthetic data, causing performance degradation when editing real videos. For example, in the top-right video frame, the robot's leg is fused with the bicycle frame. Therefore, if there is a need, future works could explore how to further stablize the finetuning, such as leveraging a prior preservation loss \cite{dreambooth} to alleviate overfitting to the dataset's domain when training steps is large.

\begin{figure*}
    \centering
    \includegraphics[width=\linewidth]{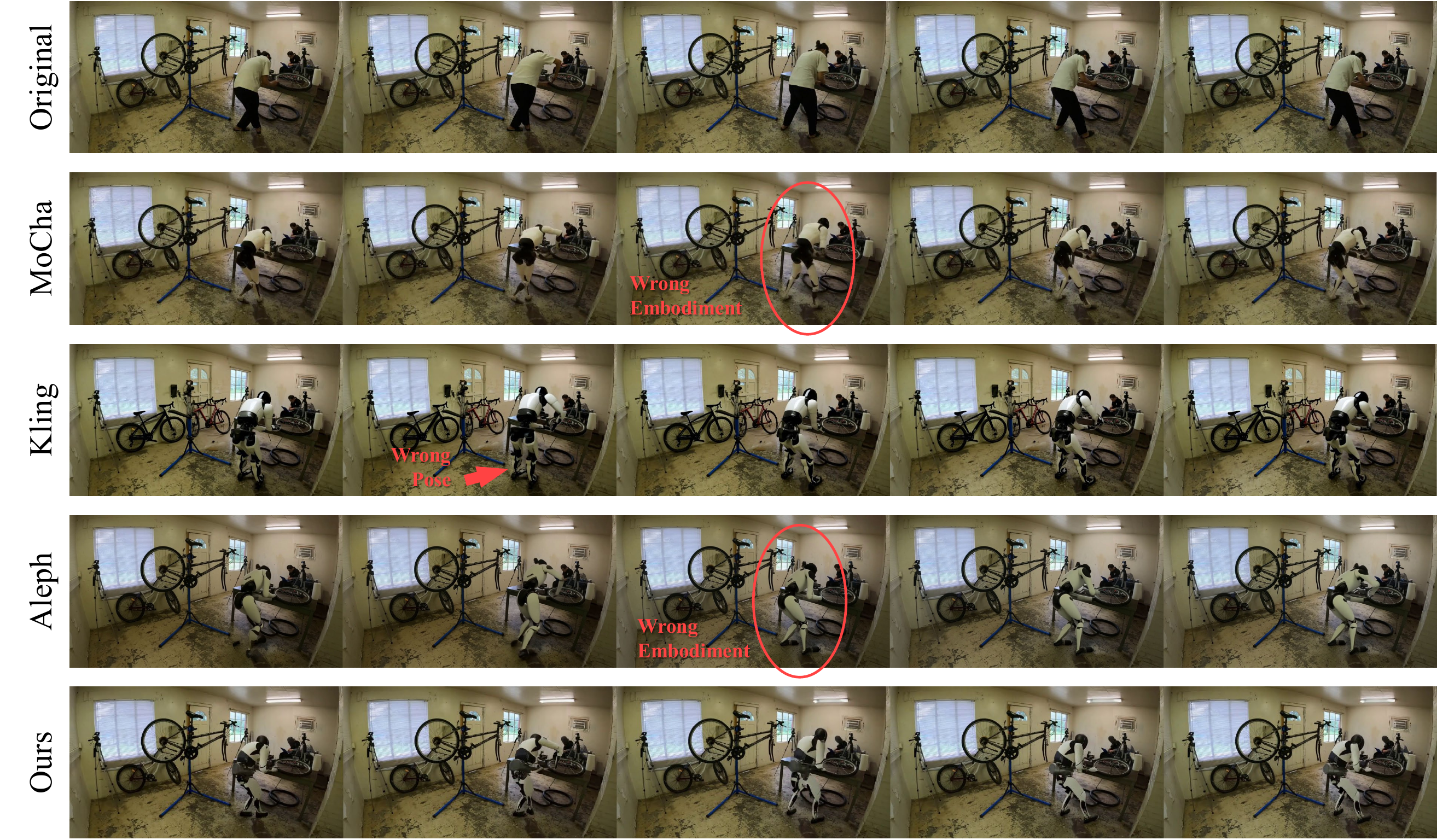}
    \caption{Qualitative comparison to baselines. The videos are sampled at frames 15, 30, 45, 60, and 75. MoCha achieves good motion consistency, but fails to generate the correct embodiment. Kling's generated embodiment is mostly correct, but the motions (such as arm and leg poses) are not aligned with the original video. For both motion consistency and embodiment correctness, Aleph's performance is between Kling and MoCha. In comparison, our method generates the most consistent motions with a correct Tesla Optimus embodiment.}
    \label{fig:appendix_baseline_1}
\end{figure*}

\begin{figure*}
    \centering
    \includegraphics[width=\linewidth]{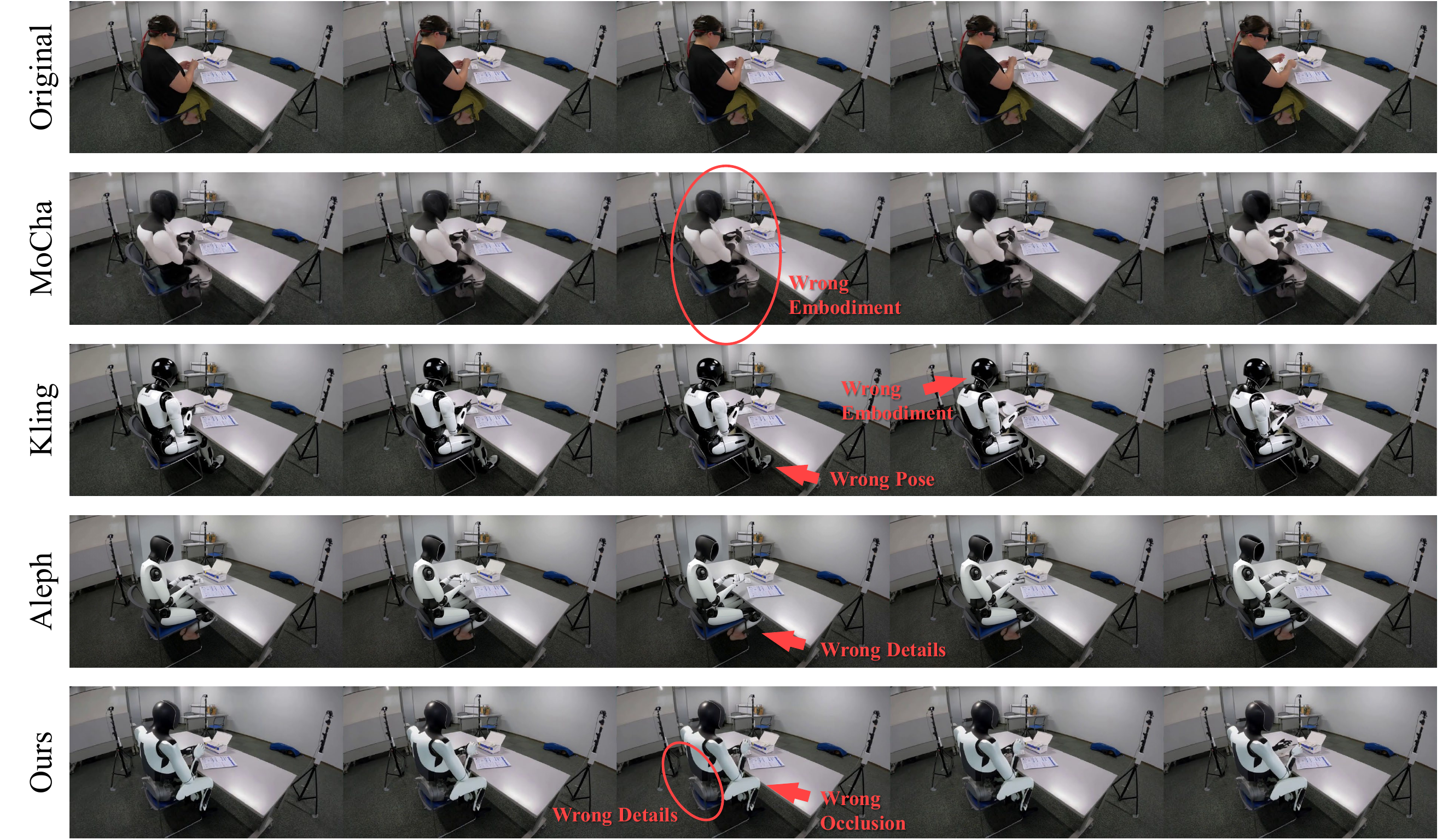}
    \caption{A failure case of our model, with comparison to baselines. The videos are sampled at frames 15, 30, 45, 60, and 75. During editing, our model may at times cause details or small objects near the robot to disappear; for example, in this video, the seat back, which is not clearly distinguished from the background color, disappears. Furthermore, for this scene, generating the leg poses under the table is also a challenge. Our method produces a slightly wrong occlusion near the knees of the humanoid, while among all methods, only Kling achieved a reasonable generation by changing the pose and location of the legs and feet under the table (although this hallucination is also undesired). This highlights that future work can focus on ensuring the correctness of details during the robotizing process.}
    \label{fig:appendix_baseline_2}
\end{figure*}

\begin{figure*}
    \centering
    \includegraphics[width=490pt]{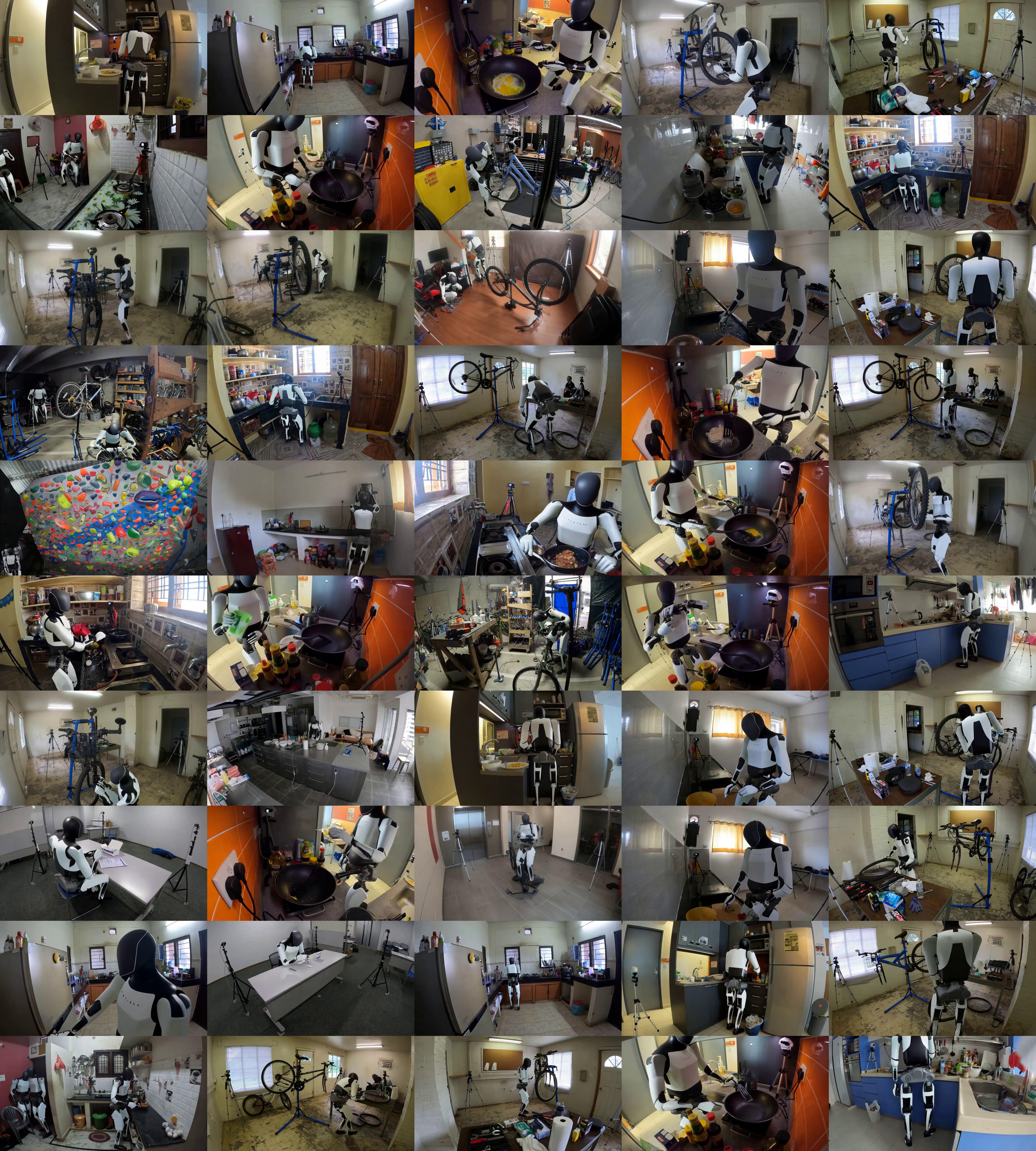}
    \caption{Visualization of all test set videos. This subset is randomly sampled, allowing for a broader inspection of our method's editing performance and the usability of the output videos.}
    \label{fig:appendix_ours_screenshots}
\end{figure*}

\section{More on Baseline Comparison}

\label{sec:appendix_baseline}

Similar to Fig.~\ref{fig:baseline_comparison}, this section provides additional qualitative visualizations comparing our method with baselines. We first present another comparison sample in Fig.~\ref{fig:appendix_baseline_1}. Following this, we present and analyze a failure case of our model in Fig.~\ref{fig:appendix_baseline_2} to highlight current limitations and provide insights for future improvements.

Additionally, we have included videos of all generation results from our method on the test set (introduced in Sec.~\ref{sec:experimental_setup}) in the supplementary materials. This subset is randomly sampled, allowing for a broader inspection of our method's editing performance and the usability of the output videos. The subset is the one shown on the project page, with screenshots in Fig. \ref{fig:appendix_ours_screenshots}.

\section{Network Inference Cost}

Editing a 480p 90-frame video takes approximately 7.5 minutes on an H200 GPU, consuming approximately 56.2GB GPU memory. 

\begin{center}
    \vspace{6pt}
    \textcolor{gray}{(End of text, figures follow)}
\end{center}


\end{document}